\documentclass[sigconf]{acmart}
\AtBeginDocument{%
  }

\copyrightyear{2026}
\acmYear{2026}
\setcopyright{cc}
\setcctype{by}
\acmConference[KDD 2026] {Proceedings of the 32nd ACM SIGKDD Conference on Knowledge Discovery and Data Mining V.2}{August 9--13, 2026}{Jeju Island, Republic of Korea.}
\acmBooktitle{Proceedings of the 32nd ACM SIGKDD Conference on Knowledge Discovery and Data Mining V.2 (KDD 2026), August 9--13, 2026, Jeju Island, Republic of Korea}
\acmISBN{979-8-4007-2259-2/2026/08}
\acmDOI{10.1145/3770855.3817565}

\acmSubmissionID{v2dtb479}

\usepackage{subcaption}
\usepackage{multirow}
\usepackage{balance}

\begin{document}


\title{PAUSE: A User-Centric Benchmark for Personal AI Assistants in Unified Service Environments}


\author{Haoyu Chen}
\affiliation{%
  \department{Electrical and Computer Engineering}
  \institution{University of Alberta}
  \city{Edmonton}
  \state{Alberta}
  \country{Canada}
}
\email{hc19@ualberta.ca}

\author{Xirui Shi}
\affiliation{%
  \department{Electrical and Computer Engineering}
  \institution{University of Alberta}
  \city{Edmonton}
  \state{Alberta}
  \country{Canada}
}
\email{xirui4@ualberta.ca}

\author{Yuyao Wang}
\affiliation{%
  \department{Electrical and Computer Engineering}
  \institution{University of Alberta}
  \city{Edmonton}
  \state{Alberta}
  \country{Canada}
}
\email{yuyao16@ualberta.ca}

\author{Jerry Chen}
\affiliation{%
  \department{Electrical and Computer Engineering}
  \institution{University of Alberta}
  \city{Edmonton}
  \state{Alberta}
  \country{Canada}
}
\email{jerry3@ualberta.ca}

\author{Di Niu}
\affiliation{%
  \department{Electrical and Computer Engineering}
  \institution{University of Alberta}
  \city{Edmonton}
  \state{Alberta}
  \country{Canada}
}
\email{dniu@ualberta.ca}

\renewcommand{\shortauthors}{Haoyu Chen, Xirui Shi, Yuyao Wang, Jerry Chen, \& Di Niu}

\begin{abstract}

Personal AI assistants are increasingly deployed as task-oriented, tool-augmented agents that operate within unified service environments to support everyday user activities. In realistic settings, such assistants must reason over persistent user state, respect user-specific configurations and permissions, and sustain long-horizon, constraint-aware interactions across multiple services. Existing benchmarks, however, often fragment service contexts or abstract away user state, limiting their ability to evaluate user-centric personal assistant behavior in realistic service settings. 
We introduce PAUSE, a user-centric benchmark for evaluating personal AI assistants in stateful, service-integrated environments. PAUSE captures core challenges of real-world assistant deployment by requiring agents to coordinate actions across heterogeneous user-owned resources while maintaining consistency with environment state and authorization constraints over multi-turn interactions. The benchmark incorporates explicit user–agent interaction via realistic user simulation, enabling evaluation beyond static tool execution.
To support principled and reproducible evaluation, PAUSE adopts a multi-regime evaluation framework aligned with task characteristics. Open-ended service management tasks are assessed using semantic and trajectory-level behavioral metrics, while constraint-intensive tasks admit deterministic, state-based verification. Benchmark results show that even state-of-the-art proprietary models fail to reach 70\% task completion on scenarios requiring stateful reasoning and configuration awareness, revealing consistent and interpretable failure patterns. Finally, we present a user-centric synthesis pipeline that enables scalable generation of coherent service environments, user configurations, and reliably annotated tasks, supporting benchmark extensibility and future research.

\end{abstract}





\maketitle
\section{Introduction}

Large language models (LLMs) are increasingly deployed as personal AI assistants, expected to support users across a wide range of everyday activities through natural-language, multi-turn interaction \cite{google_gemini_assistant,doubao_assistant,alipay_tbox}. Modern assistants are no longer limited to question answering, but are instead tool-augmented agents that retrieve data, modify persistent resources, and coordinate actions across multiple services. As a result, realistic assistant behavior requires reasoning over long-horizon interactions, persistent environment state, and user-specific constraints, including permissions, authorizations, subscriptions, and account configurations.

To evaluate these capabilities, a growing body of benchmarks has been proposed for tool-using and agentic LLMs. Early efforts emphasize API coverage and tool invocation accuracy \cite{patilberkeley, qin2023toolllm, li2023api}, while more recent benchmarks introduce multi-turn interaction, sandboxed execution, and LLM-simulated users \cite{yao2024tau, barres2025tau, lu2025toolsandbox}. In parallel, MCP-based benchmarks \cite{mo2025livemcpbench,gao2025mcp,yin2025livemcp} substantially expand scale by incorporating large numbers of real-world services and APIs, enabling stress-testing of agent planning and orchestration abilities across diverse tools.


Despite recent progress, existing benchmarks remain limited in evaluating personal AI assistants as user-facing systems. Many either focus on narrowly scoped domains or broaden coverage through MCP-style benchmarks that frame agent behavior as workflow execution over large collections of tools and services. While these settings substantially improve tool diversity and realism, they abstract away a key aspect of real-world assistant deployment: operation within a unified, user-centric service environment. In practice, assistants must reason over persistent user state and shared system configurations, with access gated by permissions, subscriptions, or prior user actions—often requiring explicit user involvement to resolve constraints. By simplifying or externalizing these factors, current benchmarks do not fully assess an AI assistant’s ability to maintain coherent state, reason about hidden configuration dependencies, and sustain user-coupled interaction over long horizons.

To address this gap, we introduce \textbf{PAUSE}, a user-centric benchmark for evaluating personal AI assistants in unified, stateful service environments. PAUSE focuses on tasks that require assistants to reason over persistent user-owned resources, respect configuration and permission constraints, and sustain long-horizon, user-coupled interaction within a single coherent system. We instantiate PAUSE in the domain of personal healthcare management, where heterogeneous data sources, external device integrations, and permission-sensitive resources naturally induce realistic and challenging environment dynamics.

Our contributions are threefold:
\begin{itemize}
\item \textbf{A holistic, user-centric benchmark}: We introduce PAUSE, a benchmark that evaluates personal AI assistants operating in unified service environments with persistent user state, shared system configurations, and permission-gated resources, moving beyond workflow-centric and tool-isolated evaluation settings.

\item \textbf{Multi-regime evaluation framework}: We design a principled evaluation strategy aligned with task characteristics, combining LLM-based semantic judgment and trajectory-level behavioral overlap for open-ended data and log tracking tasks, with deterministic state-based verification for constraint-intensive shopping tasks, enabling reliable comparison and interpretable diagnosis of agent behavior.

\item \textbf{Scalable user-centric synthesis pipeline}: We present a multi-agent task generation pipeline that produces coherent environment states, realistic agent–user interaction trajectories, and target-aligned annotations at scale, supporting benchmark extensibility and downstream applications such as model analysis and distillation.

\end{itemize}

Extensive evaluation reveals that even state-of-the-art proprietary models fail to reach 70\% task completion on tasks requiring stateful reasoning and configuration awareness, with errors exhibiting consistent and interpretable structure across task types. We have uploaded the source code of the benchmark to \url{https://github.com/hyc481/PAUSE}.

\section{Related Work}

\begin{table*}[t]
\caption{\textbf{Comparison of PAUSE with representative benchmarks.}
\textbf{Stateful Init}: whether each task starts from a task-specific environment state.
\textbf{Holistic System}: whether the benchmark models a unified service environment.
\textbf{Execution}: how tool calls are executed.
\textbf{User Interaction}: whether explicit user--agent multi-turn interaction is included.
\textbf{Evaluation Regime}: rule-based verification, LLM-as-judge, or hybrid.
\textbf{Pipeline}: whether an end-to-end pipeline is provided, including task generation and multi-turn tool-calling reference trajectory generation.}
\label{tab:benchmark_comparison}
\centering
\small
\setlength{\tabcolsep}{5.5pt}
\renewcommand{\arraystretch}{1.15}
\begin{tabular}{lcccccc}
\toprule
\textbf{Benchmark} &
\textbf{Stateful Init.} &
\textbf{Holistic System} &
\textbf{Execution} &
\textbf{User Interaction} &
\textbf{Evaluation Regime} &
\textbf{Pipeline} \\
\midrule
BFCL \cite{patilberkeley}                     
& \textcolor{red}{$\times$}  
& \textcolor{red}{$\times$}  
& Sandbox (multi-turn) 
& \textcolor{red}{$\times$}  
& Rule-based   
& \textcolor{red}{$\times$}  \\

ToolBench \cite{qin2023toolllm}              
& \textcolor{red}{$\times$}  
& \textcolor{red}{$\times$}  
& RapidAPI \cite{rapidapi_hub}             
& \textcolor{red}{$\times$}  
& LLM-as-judge 
& \textcolor{green}{$\checkmark$} \\

StableToolBench \cite{guo2024stabletoolbench}
& \textcolor{red}{$\times$}  
& \textcolor{red}{$\times$}  
& LLM simulation       
& \textcolor{red}{$\times$}  
& LLM-as-judge 
& \textcolor{green}{$\checkmark$} \\

ToolSandbox \cite{lu2025toolsandbox}         
& \textcolor{green}{$\checkmark$} 
& \textcolor{red}{$\times$}  
& Sandbox              
& \textcolor{green}{$\checkmark$} 
& Rule-based   
& \textcolor{red}{$\times$}  \\

ACEBench \cite{chen2025acebench}             
& \textcolor{red}{$\times$}  
& \textcolor{red}{$\times$}  
& Sandbox              
& \textcolor{green}{$\checkmark$} 
& Rule-based   
& \textcolor{red}{$\times$}  \\

LiveMCPBench \cite{mo2025livemcpbench}       
& \textcolor{red}{$\times$}  
& \textcolor{red}{$\times$}  
& MCP servers          
& \textcolor{red}{$\times$}  
& LLM-as-judge 
& \textcolor{red}{$\times$}  \\

$\tau$-Bench \cite{yao2024tau}               
& \textcolor{red}{$\times$}  
& \textcolor{red}{$\times$}  
& Sandbox              
& \textcolor{green}{$\checkmark$} 
& Rule-based   
& \textcolor{red}{$\times$}  \\

$\tau^2$-Bench \cite{barres2025tau}          
& \textcolor{green}{$\checkmark$} 
& \textcolor{red}{$\times$}  
& Sandbox              
& \textcolor{green}{$\checkmark$} 
& Rule-based   
& \textcolor{red}{$\times$}  \\

\textbf{PAUSE (Ours)}                        
& \textcolor{green}{$\checkmark$} 
& \textcolor{green}{$\checkmark$} 
& Sandbox              
& \textcolor{green}{$\checkmark$} 
& Rule + LLM   
& \textcolor{green}{$\checkmark$} \\

\bottomrule
\end{tabular}
\end{table*}

We briefly review related work on benchmarks and evaluation frameworks for language agents and tool-using assistants. A comprehensive comparison between PAUSE and representative benchmarks is presented in Table~\ref{tab:benchmark_comparison}.

\textbf{API-Oriented Tool-Usage Benchmarks.}
Benchmarks for tool-using LLMs have been widely studied, with early work primarily emphasizing large-scale tool coverage by curating extensive API collections and test cases across diverse usage patterns \cite{patilberkeley,shen2024shortcutsbench,tang2023toolalpaca,patil2024gorilla,li2023api,qin2023toolllm,guo2024stabletoolbench,xu2023tool}. Representative efforts such as ToolBench \cite{qin2023toolllm} and ShortcutsBench \cite{shen2024shortcutsbench} construct realistic tool inventories from public API hubs \cite{rapidapi_hub,apple_shortcuts_app}, while BFCL \cite{patilberkeley} unifies multiple API sources and evaluation protocols and has served as a standard benchmark for early tool-usage evaluation. To improve scalability and reproducibility, several works further leverage LLMs to synthesize APIs and tool responses \cite{tang2023toolalpaca,li2023api,guo2024stabletoolbench}.
This emphasis on maximizing tool coverage—often at the expense of explicit system design and sandboxed execution—reduces evaluation largely to tool selection and parameter filling, leaving stateful decision-making and long-horizon planning underexplored. Other benchmarks instead focus on specific facets of tool-using agents, including user preference perception and alignment \cite{qian2025userbench}, failure awareness under underspecified requests or unavailable tools \cite{trevino2025benchmarking}, workflow-following with demonstrations FlowBench \cite{xiao2024flowbench}, and multi-turn planning and interaction \cite{chakraborty2025t1}.

\textbf{User-agent Interplay and Stateful Sandboxes.} 
More recently, multi-turn user–assistant interaction with stateful sandbox execution has emerged as a more discriminative paradigm for evaluating agent capabilities \cite{lu2025toolsandbox,yao2024tau,barres2025tau,chen2025acebench}. These benchmarks simulate agent–user interplay via LLM-based user roleplay within sandboxed environments that preserve system state and produce structured tool feedback. For instance, ToolSandbox \cite{lu2025toolsandbox} evaluates agents under diverse initialized world states, making task success inherently state-dependent. $\tau^2$-Bench \cite{barres2025tau} further examines interactive state modification by user under domain-specific constraints, but is confined to a single vertical, limiting its ability to reflect holistic assistant behavior across services. We further extend these work to a coherent, unified environment spanning heterogeneous personal service tasks, with explicit modeling of user-centric configurations and constraints for realistic personal assistant evaluation.

\textbf{Evaluation Paradigms for Tool-Using Agents.} 
Existing agent benchmarks primarily adopt two evaluation paradigms: rule-based verification and LLM-as-judge. Early rule-based methods, such as BFCL \cite{patilberkeley}, evaluate tool usage via deterministic tool name and parameter matching, often lacking tolerance to semantic variation.
Benchmarks with sandboxed backends, including $\tau$-Bench, $\tau^2$-Bench, and ToolSandbox \cite{yao2024tau, barres2025tau, lu2025toolsandbox}, instead rely on state alignment between agent actions and environment transitions. While precise, such evaluation is inherently restricted to narrowly defined, state-verifiable tasks and does not generalize to semantically flexible behaviors.
In contrast, LLM-as-judge evaluation—used for semantic validation, trajectory comparison, and behavioral consistency assessment—is widely adopted for open-ended tool use. ToolBench \cite{qin2023toolllm} applies LLM-based judgment for both data curation and trajectory evaluation, while LiveMCP-101 \cite{yin2025livemcp} and LiveMCPBench \cite{mo2025livemcpbench} extend this paradigm with execution-plan alignment and structured, keypoint-grounded judging respectively.
Motivated by PAUSE’s task diversity, we adopt a hybrid evaluation strategy: semantically flexible data \& log tracking tasks are assessed via LLM-based judgment, whereas state- and constraint-intensive shopping tasks admit deterministic state-based verification.

\textbf{Pipelines for Tool-Usage Data Synthesis.} 
Pipelines for tool-usage benchmarks are primarily designed to either synthesize training data or construct annotated benchmark tasks. Representative approaches such as APIGen \cite{liu2024apigen}, ToolBench \cite{qin2023toolllm}, and ToolACE \cite{liu2024toolace} combine powerful LLMs with rule-based and semantic validation to distill high-quality tool-calling trajectories, with subsequent extensions exploring iterative self-refinement \cite{zeng2025toolace_r}, graph-based synthesis \cite{yin2025magnet}, and single-pass multi-turn trajectory generation \cite{zeng2025toolace_mt}.
However, these pipelines largely target BFCL-style \cite{patilberkeley} benchmarks with large but shallow tool pools, and therefore struggle to synthesize long-horizon, state-dependent interactions requiring coherent environment dynamics. In contrast, APIGen-MT \cite{prabhakar2025apigen} builds on $\tau$-bench \cite{yao2024tau} and proposes an agentic synthesis pipeline that generates multi-round trajectories via simulated human–agent interplay, grounded in executable backend environments and verified through real execution. Relatedly, $\tau^2$-Bench \cite{barres2025tau} proposes a pipeline for synthesizing tasks with annotated state specifications, but does not generate interactive agent–user trajectories as reference. Inspired by these works, we develop a synthesis pipeline that generates annotated tasks from real execution with stateful tool calls, while additionally constructing structured task targets for task availability validation and structured evaluation references to support downstream LLM-as-judge evaluation.

\textbf{MCP Benchmarks}. Recent benchmarks further explore tool usage under the Model Context Protocol (MCP) setting \cite{anthropic2024modelcontextprotocol}, emphasizing large-scale tool orchestration and workflow execution across heterogeneous services.
MCP-RADAR \cite{gao2025mcp} introduces a multi-dimensional evaluation framework for MCP-enabled tool use with objective metrics over real and high-fidelity simulated tools.
LiveMCPBench \cite{mo2025livemcpbench} and LiveMCP-101 \cite{yin2025livemcp} benchmark multi-step, real-world tasks over live MCP servers, adopting LLM-as-judge to handle dynamic tool responses.
TOOLDECATHLON \cite{li2025tool} further stresses scalability by evaluating long-horizon, cross-application tasks over hundreds of tools.
Despite their scale and realism, these benchmarks primarily frame agent behavior as workflow orchestration over large tool ensembles, and differ fundamentally from settings that require agents to operate within a coherent, holistic, user-centric service environment with persistent user state, configuration, and permissions.

\section{Benchmarking Personal AI Assistants in Holistic, User-centric Service Environments}
\subsection{System Overview}
\begin{figure}[htbp]
    \centering
    \includegraphics[width=\linewidth]{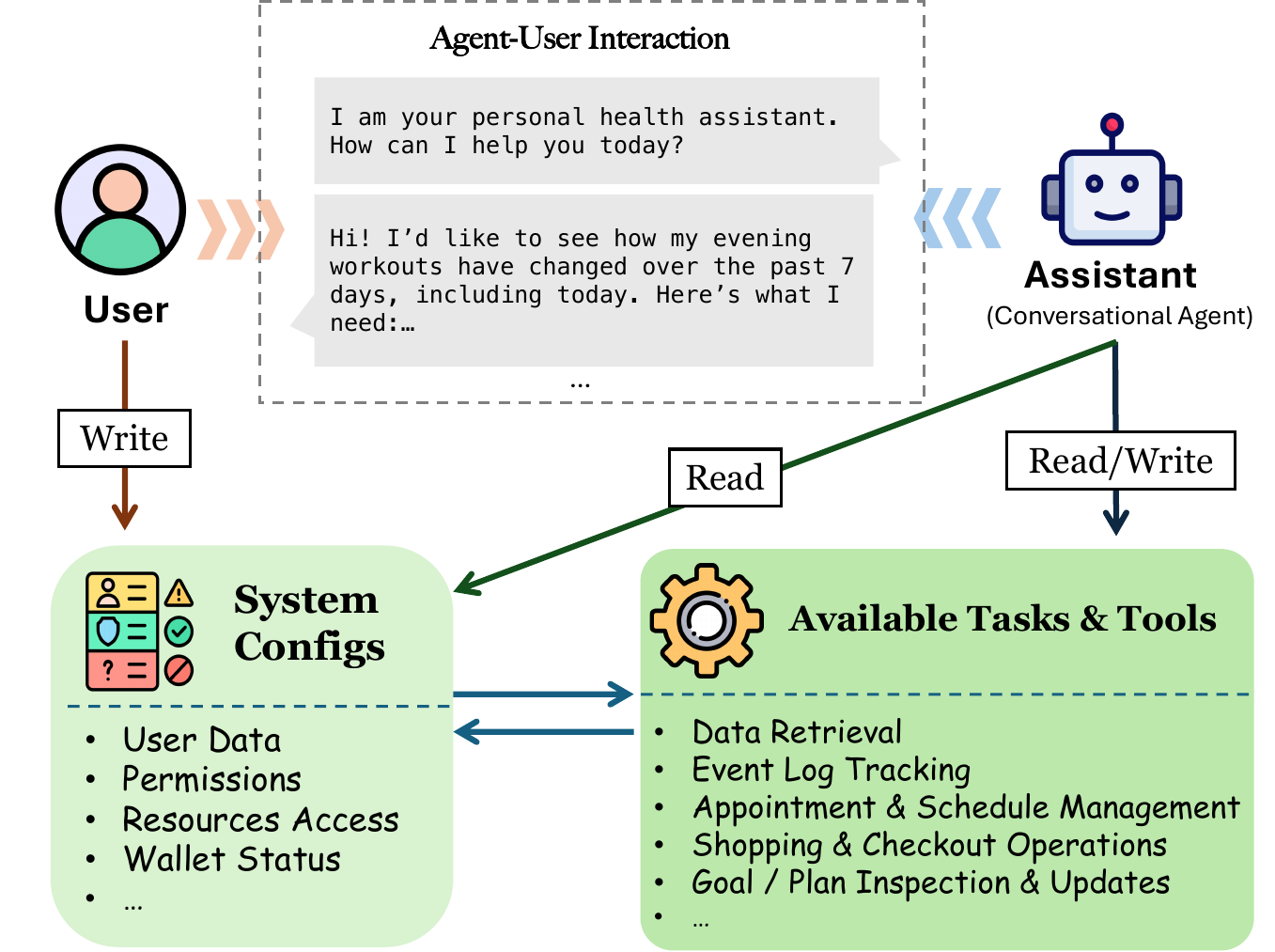}
    \caption{System Overview. PAUSE simulates a user-centric personal service environment where personal AI assistants operate under user permissions and system configurations.}
    \Description{A high-level diagram illustrating the PAUSE system architecture. The figure shows a user interacting with a personal AI assistant, which operates within a simulated personal service environment.}
    \label{fig:env_overview}
\end{figure}

PAUSE is a benchmark that simulates a unified personal service environment for evaluating natural-language assistants under realistic, user-specific constraints. As illustrated in Figure~\ref{fig:env_overview}, the assistant interacts with the user through natural language and executes tasks by invoking tools with read and write access. While the assistant can read from and write to task-related resources, it can only read critical system configurations that gate permissions, resources, and account states; such configurations can be modified only through explicit user actions. This user-centric and holistic setting is designed to mirror the security and authorization constraints found in real-world personal service systems.

\textbf{User-centric.} The environment is user-centric in that all tasks are grounded in persistent, user-owned resources and permissions, covering a wide range of everyday activities such as health data tracking, profile and account inspection, scheduling, shopping, and plan management. Access to key resources is explicitly controlled by the user, requiring assistants to reason about authorization states and request permissions when needed rather than assuming full access.

\textbf{Holistic system configuration.} The environment is holistic because all services operate under a shared, persistent system configuration, including permission settings, resource availability, wallet and subscription status, and other hidden platform states. A single task often involves navigating multiple services and triggering different permissions or state transitions, where actions in one service directly affect subsequent interactions across others.

A more detailed system specification can be found in Appendix~\ref{sec:system_specification}. Our simulated user-centric, holistic environment is challenging due to the following features:
\begin{itemize}
  \item \textbf{Multi-granularity temporal data.}  
  Data tracking tasks involve aggregated data at different temporal resolutions, requiring the assistant to perform temporal reasoning and select appropriate time spans to satisfy user requests.
  \item \textbf{Large and diverse tool space.}  
  The environment exposes 50 assistant tools and 7 user tools within a shared system context, inducing a diverse task space that involves resource navigation and long-context reasoning.
  \item \textbf{Implicit system configuration.}  
  System configurations are largely not directly visible to the assistant and must be inferred from tool responses; the assistant must interact with the user to satisfy configuration or permission requirements.
  \item \textbf{Hidden and gated tools.}  
  Certain tools are inaccessible until system configuration requirements are satisfied, such as raw pre-aggregation data or resources from external connected applications (e.g., healthcare providers), simulating gated access in real-world service systems.
\end{itemize}

\subsection{Pipeline Setup}

We model PAUSE as a partially observed user–assistant interaction process over an environment state. 
At step \(t\), the latent environment state is
\begin{displaymath}
s_t = (D_u, I_u, C_u),
\end{displaymath}
where \(D_u\) denotes user data, \(I_u\) denotes user profile and account information, and \(C_u\) denotes system and user configurations, including permissions, wallet state, subscriptions, and access to hidden service APIs and external resources.
Neither the user nor the assistant directly observes \(s_t\). Instead, interaction is mediated through observations derived from the state and recent interactions. We denote the observation available to the assistant as
\begin{displaymath}
o_t = \Omega(s_t, h_t),
\end{displaymath}
where \(h_t\) is the dialogue and tool-call history. Specifically, \(o_t\) consists of the natural-language dialogues and tool responses, forming a partial and compressed view of \(s_t\).
The interaction involves two action spaces: assistant actions \(a_t \in \mathcal{A}\) and user actions \(u_t \in \mathcal{U}\). We extend the dual-control environment setting introduced in $\tau^2$-Bench \cite{barres2025tau} to simulate user control over critical system states and configurations in our benchmark. Interaction proceeds as an alternating sequence beginning with a user instruction:
\begin{displaymath}
x_t^{u} \;\rightarrow\; a_t \;\rightarrow\; o_t \;\rightarrow\; y_t^{a} \;\rightarrow\; u_t \;\rightarrow\; x_{t+1}^{u} \;\rightarrow\; \dots
\end{displaymath}
where \(x_t^{u}\) and \(y_t^{a}\) denote natural-language responses from user and assistant respectively.

\begin{figure*}[htbp]
    \centering
    \includegraphics[width=\linewidth]{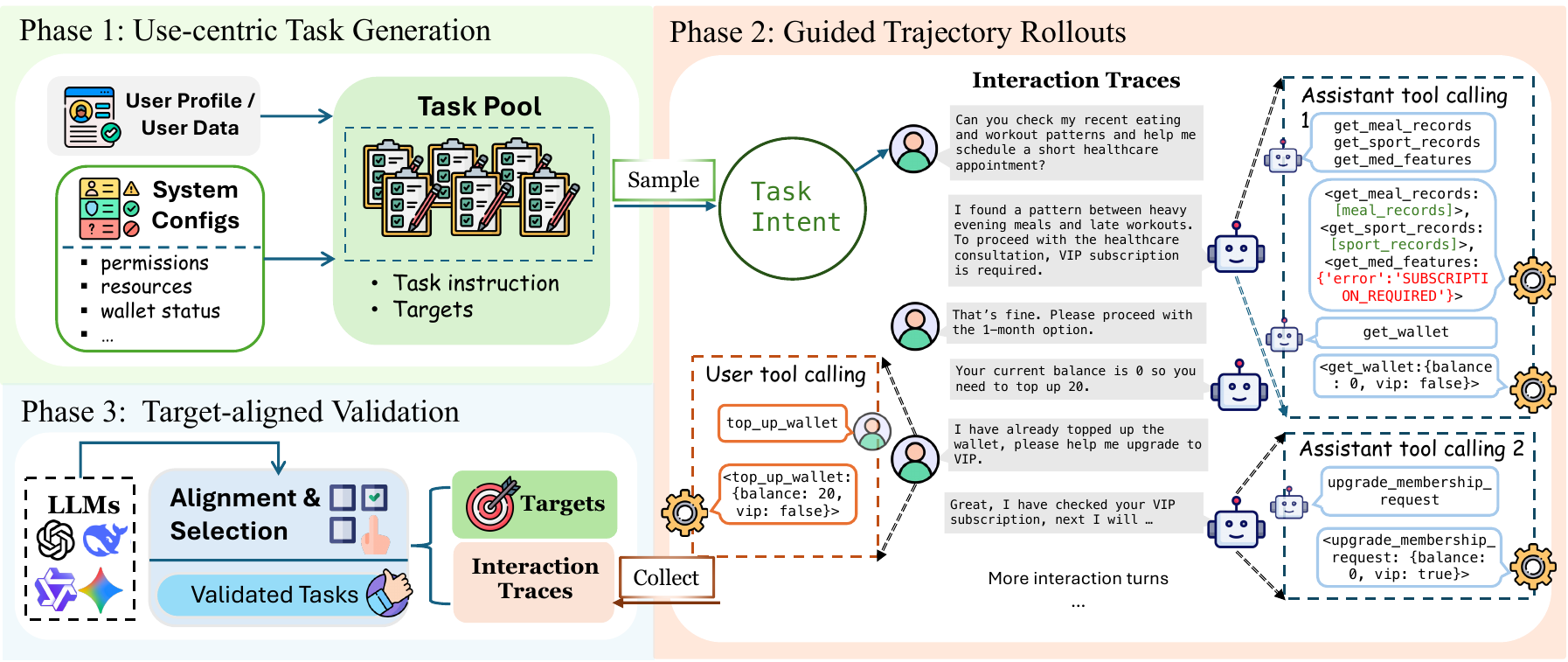}
    \caption{Overview of the PAUSE pipeline, which consists of three stages: (1) user-centric task generation, (2) guided trajectory rollouts, and (3) target-aligned evaluation.}
    \label{fig:pipeline}
    \Description{An overview of the PAUSE pipeline composed of three sequential stages. The first stage generates user-centric tasks based on simulated user profiles and environment states. The second stage executes guided trajectory rollouts where an assistant interacts with the environment to complete the tasks. The third stage performs target-aligned evaluation by assessing the resulting trajectories against predefined task targets.}
\end{figure*}

\textbf{User-centric Task Generation.}
As demonstrated in Figure~\ref{fig:pipeline}, our pipeline starts with an explicit user–system configuration generation.
Instead of relying on stochastic LLM-generated states, we achieve both diversity and structural completeness by grounding each task in synthetic, template-sampled data that emulates concrete, user-owned resources.
Crucially, different task templates are associated with different data injection schemas.
 Formally, the initial configuration is constructed as
\begin{displaymath}
    s_{t_0} = \mathcal{I}(\mathcal{D}, \mathcal{D}_k),
\end{displaymath}

where $\mathcal{D}$ denotes base structured user data sampled from generic templates, and $\mathcal{D}_k$ denotes task-specific injected data determined by template $k$. Injected data include user activity records (e.g., meal and activity logs) as well as wearable device data \cite{fitbit_kaggle}. Each task template is defined as
\begin{displaymath}
\mathcal{T}_k = \{\mathcal{E}_k,\; P_k^{\text{task}},\; P_k^{\text{rollout}},\; \mathcal{O}_k,\; \mathcal{D}_k\},
\end{displaymath}
where $\mathcal{E}_k$ is a pool of few-shot examples, $P_k^{\text{task}}$ is the task-generation prompt, $P_k^{\text{rollout}}$ specifies rollout assistant guidance, and $\mathcal{O}_k$ is an observation function that governs information exposure.

Given $s_{t_0}$, an LLM first produces a high-level, user-facing description $\tilde{s}_{t_0} = \mathrm{LLM}(s_{t_0})$.
The observation function then maps the underlying configuration to observable signals $o_{t_0} = \mathcal{O}_k(s_{t_0})$.
Conditioned on the rendered description, observable signals, and a sampled subset of few-shot examples $\mathcal{E}_k' \subseteq \mathcal{E}_k$, the LLM jointly generates a task instruction and an explicit set of target conditions:
\begin{displaymath}
(q, \boldsymbol{\tau}) = \mathrm{LLM}(\tilde{s}_{t_0}, o_{t_0}, \mathcal{E}_k'; P_k^{\text{task}}),
\end{displaymath}
where the task instruction is provided to a user agent during trajectory rollouts to act as the user and issue task commands, and $\boldsymbol{\tau} = [\tau_1, \ldots, \tau_m]$ denotes a set of verifiable target conditions.

\textbf{Guided Trajectory Rollouts.}
Given a generated task $(q, \boldsymbol{\tau})$ and the corresponding initial configuration $s_{t_0}$, we obtain candidate interaction trajectories through real execution.
Specifically, rollouts are conducted under a rollout prompt $P_k^{\text{rollout}}$, which specifies desired assistant behavior and serves as an oracle policy for the benchmark. Formally, $P_k^{\text{rollout}}$ induces a constrained assistant policy:
\begin{displaymath}
P_k^{\text{rollout}} \mapsto \mathcal{A}_k \subseteq \mathcal{A},
\end{displaymath}
Under this guidance, multiple state-of-the-art LLMs interact with the system and user to collect real-execution rollouts, producing a set of candidate trajectories
\begin{displaymath}
\{\xi_i\}_{i=1}^N, \quad \xi_i = (x_{t_0}^{u}, a_{t_0}, o_{t_0}, y_{t_0}^{a}, u_{t_0}, \ldots).
\end{displaymath}

\textbf{Trajectory Annotation and Target Alignment.}
Candidate trajectories are subsequently filtered and annotated by an LLM committee.
Each trajectory first undergoes a trajectory-level pass/fail screening, and infeasible candidates are discarded.
For the remaining trajectories, the committee produces concise summaries and selects a preferred trajectory $\xi^\ast$; when no candidate is satisfactory, additional rollouts are triggered with feedback-guided refinement.
The selected trajectory is then explicitly aligned against the predefined target set $\boldsymbol{\tau}$.
Tasks for which no trajectory satisfies all target conditions are removed from the benchmark.
By introducing explicit and verifiable targets $\boldsymbol{\tau}$, PAUSE guarantees the existence of feasible reference trajectories and provides a structured scoring signal for downstream evaluation in the absence of canonical solutions. Appendix~\ref{sec:prompts_examples} provides representative few-shot examples used in the pipeline generation process.

\subsection{Multi-regime Evaluation Framework}
PAUSE adopts a multi-regime evaluation framework tailored to task characteristics. 
For data and log tracking tasks, which lack canonical solutions, we evaluate task completion using an LLM-as-judge guided by explicit targets $\boldsymbol{\tau}$, which serve as the primary scoring signal. 
To complement target-based judgment and improve evaluation reliability, we additionally compute trajectory-level overlap metrics between a selected reference trajectory and the model-generated trajectory.

Let $M$ and $\hat{M}$ denote the reference and evaluated trajectories, respectively. 
Each trajectory is represented as a multiset of tool-call elements 
$e=(\texttt{tool\_name},\texttt{arguments})$, with counts $c_M(e)$ and $c_{\hat{M}}(e)$. 
Precision, recall, and F1 are computed as:
\begin{displaymath}
\begin{aligned}
\mathrm{P} &=
\frac{\sum_e \min(c_M(e), c_{\hat{M}}(e))}
     {\sum_e c_{\hat{M}}(e)}, \\
\mathrm{R} &=
\frac{\sum_e \min(c_M(e), c_{\hat{M}}(e))}
     {\sum_e c_M(e)}, \\
\mathrm{F1} &=
\frac{2\mathrm{P}\mathrm{R}}{\mathrm{P}+\mathrm{R}} .
\end{aligned}
\end{displaymath}

For target-based evaluation, given target set 
$\boldsymbol{\tau}=\{\tau_1,\ldots,\tau_m\}$, target completion is defined as:
\begin{displaymath}
\mathrm{TC}=\frac{1}{m}\sum_{j=1}^{m}\mathbb{I}(\tau_j),
\end{displaymath}
and final results report the average $\mathrm{TC}$ across tasks.

For shopping tasks, which admit explicit state representations and deterministic constraint verification, evaluation is performed via state-based checking without relying on reference trajectories or target-based scoring. Each task is decomposed into a set of verifiable targets, including target product identification, quantity and size selection, voucher usage, and budget feasibility. The final task score is computed by aggregating satisfaction over these targets. 

\section{Experiments}
\subsection{Experiment Setup}

All tasks in PAUSE are collected through the proposed pipeline.
We use Gemini-3-Flash \cite{google_gemini3_flash}, GPT-5 \cite{openai_gpt5}, and GPT-5-Mini \cite{openai_gpt5_mini} as the primary agents for both trajectory collection and annotation, with Gemini-3-Flash additionally serving as the user agent during roleplay.
The pipeline generates over 300 candidate tasks across domains, from which we select 180 tasks for the final evaluation set after filtering for clarity, executability, and target alignment.
The final test set consists of three categories:
\textit{(i)} 63 data \& log tracking (\emph{easy}) tasks,
\textit{(ii)} 57 data \& log tracking (\emph{hard}) tasks, and
\textit{(iii)} 60 shopping tasks.

Easy tasks focus on direct data and log retrieval.
Hard tasks extend beyond retrieval to multi-step service execution, such as scheduling clinical appointments or upgrading to VIP to unlock subscription-based services. These tasks often involve initially unavailable data or services due to unmet system configurations (e.g., unconnected data sources or missing permissions), requiring the assistant to reason about system constraints, interact with the user, and trigger prerequisite operations before completion.
Table~\ref{tab:tracking_task_stats} reflects this key distinction: hard tasks exhibit longer interactions and higher tool-call frequency, and notably are the only setting in which user tool calls appear, highlighting the need for user-mediated system-level actions. Shopping tasks require the assistant to identify an optimal product that satisfies multiple user-specified constraints, including price, quantity, size, discounts, and nutritional attributes. This setting fundamentally constitutes a long-context multi-constraint optimization problem. The assistant must also correctly manage stateful interactions, such as maintaining the shopping cart, handling budget constraints, applying vouchers, and coordinating user interactions for critical state transitions (e.g., wallet top-up or checkout authorization).

We evaluate a diverse set of state-of-the-art large language models, covering both proprietary and open-source systems.
The proprietary models include GPT-5 and GPT-5 mini \cite{openai_gpt5,openai_gpt5_mini}, GPT-4.1 and GPT-4.1 mini \cite{openai_gpt4_1,openai_gpt4_1_mini}, as well as Gemini-3-Pro, Gemini-2.5-Pro and Gemini-3-Flash \cite{google_gemini3_pro, google_gemini2_5_pro, google_gemini3_flash}.
For open-source models, we consider DeepSeek-V3.2 with and without thinking mode \cite{liu2025deepseek}.

\begin{table}[t]
\centering
\small
\caption{Statistics of data \& log tracking tasks. Avg. Rounds denotes the average number of dialogue turns. Assistant TC and User TC indicate the average number of tool calls made by the assistant and the user simulator, respectively.}
\label{tab:tracking_task_stats}
\begin{tabular}{lccc}
\toprule
\textbf{Task Type} & \textbf{Avg. Rounds} & \textbf{Assistant TC} & \textbf{User TC} \\
\midrule
Data \& Log Tracking (Easy) & 2.11 & 12.85 & 0.00 \\
Data \& Log Tracking (Hard) & 5.23 & 22.07 & 3.28 \\
\bottomrule
\end{tabular}
\end{table}

\subsection{Main Results}
We analyze model behavior from two complementary perspectives: interaction complexity and task performance. 
Tables~\ref{tab:easy_calls_rounds} and~\ref{tab:hard_calls_rounds} summarize the average number of assistant tool calls, user tool calls, and dialogue rounds for data \& log tracking tasks across all evaluated models. Compared to easy tasks, data \& log tracking (\emph{hard}) tasks consistently require more dialogue rounds and tool invocations across all evaluated models, reflecting their increased interaction complexity. Hard tasks require assistants to sustain longer trajectories while coordinating with user-side actions and evolving environment states.

We adopt Gemini-3-Flash as the default evaluator model. As shown in Tables~\ref{tab:tracking_easy} and~\ref{tab:tracking_hard}, data \& log tracking tasks exhibit clear performance stratification across model tiers. On easy tasks, top-tier proprietary models (e.g., Gemini-3-Flash/Pro, GPT-5, GPT-5-Mini) achieve near-saturated performance with only marginal differences, while DeepSeek-V3.2-Thinking shows a moderate gap and weaker models lag far behind. Performance degrades substantially on hard tasks that require system configuration reasoning and environment state tracking. Even frontier models experience notable drops, revealing limited robustness in handling implicit system constraints beyond direct tool invocation. This contrast further widens the gap between model tiers, with weaker models consistently underperforming across both settings and struggling most severely under increased statefulness and reasoning demands.

As shown in Table~\ref{tab: shopping}, shopping tasks yield a similar model ranking to data \& log tracking tasks, with proprietary models leading overall. DeepSeek-V3.2-Thinking demonstrates more competitive performance on shopping tasks, approaching state-of-the-art proprietary models. 

\begin{table}[!htbp]
\centering
\small
\caption{Average assistant tool calls, user tool calls, and dialogue rounds on data \& log tracking (easy) tasks.}
\label{tab:easy_calls_rounds}
\begin{tabular}{lccc}
\toprule
Model & Asst. Calls & User Calls & Rounds \\
\midrule
Gemini-3-Flash & 13.55 & 0.00 & 2.08 \\
Gemini-3-Pro   & 10.58 & 0.00 & 2.11 \\
Gemini-2.5-Pro & 15.31 & 0.03 & 2.22 \\
GPT-5          & 12.04 & 0.00 & 2.16 \\
GPT-5-Mini     & 12.28 & 0.00 & 2.22 \\
GPT-4.1        & 11.12 & 0.00 & 2.32 \\
GPT-4.1-Mini   & 9.50  & 0.00 & 2.65 \\
DeepSeek-V3.2-Thinking    & 12.77 & 0.00 & 3.41 \\
DeepSeek-V3.2  & 14.58 & 0.00 & 2.95 \\
\bottomrule
\end{tabular}
\end{table}

\begin{table}[!htbp]
\centering
\small
\caption{Average assistant tool calls, user tool calls, and dialogue rounds on data \& log tracking (hard) tasks.}
\label{tab:hard_calls_rounds}
\begin{tabular}{lccc}
\toprule
Model & Asst. Calls & User Calls & Rounds \\
\midrule
Gemini-3-Flash & 24.56 & 3.40 & 4.13 \\
Gemini-3-Pro   & 17.63 & 2.86 & 4.26 \\
Gemini-2.5-Pro & 15.21 & 2.58 & 5.33 \\
GPT-5          & 33.00 & 3.17 & 5.21 \\
GPT-5-Mini     & 35.93 & 3.18 & 5.88 \\
GPT-4.1        & 14.90 & 3.05 & 6.86 \\
GPT-4.1-Mini   & 13.56 & 2.40 & 6.21 \\
DeepSeek-V3.2-Thinking    & 25.60 & 3.28 & 5.14 \\
DeepSeek-V3.2  & 27.05 & 3.09 & 4.79 \\
\bottomrule
\end{tabular}
\end{table}

\begin{table}[t]
\centering
\caption{Performance on data \& log tracking tasks (Easy). We report task completion (TC), average targets achieved (TA), and trajectory overlap precision/recall/F1 (Pre/Rec/F1).}
\label{tab:tracking_easy}
\small
\begin{tabular}{lccccc}
\toprule
Model & TC. & TA. & Pre & Rec & F1 \\
\midrule
Gemini-3-Flash         & 85.72\% & 95.98\% & 0.841 & 0.796 & 0.796 \\
Gemini-3-Pro           & 92.06\% & 98.51\% & \textbf{0.849} & 0.869 & 0.844 \\
Gemini-2.5-Pro         & 66.70\% & 90.23\% & 0.733 & 0.751 & 0.711 \\
GPT-5                  & \textbf{95.26\%} & \textbf{98.85\%} & 0.847 & \textbf{0.880} & 0.856 \\
GPT-5-Mini             & 92.07\% & 98.37\% & 0.895 & 0.876 & \textbf{0.880} \\
GPT-4.1                & 33.34\% & 71.18\% & 0.669 & 0.584 & 0.588 \\
GPT-4.1-Mini           & 28.56\% & 62.41\% & 0.648 & 0.564 & 0.582 \\
DeepSeek-V3.2-Thinking & 69.86\% & 85.00\% & 0.718 & 0.744 & 0.720 \\
DeepSeek-V3.2          & 47.57\% & 72.14\% & 0.560 & 0.623 & 0.578 \\
\bottomrule
\end{tabular}
\end{table}

\begin{table}[t]
\centering
\caption{Performance on data \& log tracking tasks (Hard).}
\label{tab:tracking_hard}
\small
\begin{tabular}{lccccc}
\toprule
Model & TC. & TA. & Pre & Rec & F1 \\
\midrule
Gemini-3-Flash         & \textbf{59.12\%} & \textbf{77.48\%} & \textbf{0.584} & 0.492 & \textbf{0.517} \\
Gemini-3-Pro           & 48.39\% & 77.25\% & 0.511 & 0.427 & 0.439 \\
Gemini-2.5-Pro         & 19.33\% & 57.76\% & 0.516 & 0.343 & 0.379 \\
GPT-5                  & 47.34\% & 72.96\% & 0.494 & \textbf{0.555} & 0.479 \\
GPT-5-Mini             & 43.84\% & 66.96\% & 0.439 & 0.495 & 0.406 \\
GPT-4.1                & 17.56\% & 57.00\% & 0.401 & 0.279 & 0.303 \\
GPT-4.1-Mini           & 10.53\% & 38.81\% & 0.436 & 0.293 & 0.331 \\
DeepSeek-V3.2-Thinking & 35.11\% & 55.71\% & 0.297 & 0.326 & 0.296 \\
DeepSeek-V3.2          & 14.06\% & 40.02\% & 0.257 & 0.245 & 0.231 \\
\bottomrule
\end{tabular}
\end{table}

\begin{table}[t]
\centering
\small
\caption{Performance on shopping tasks across models. PID denotes the average product purchase completion rate. Qty\_Size measures whether the assistant selects the optimal quantity and size; Voucher indicates the proportion of purchase completed with optimal voucher selection; Balance denotes the proportion of tasks completed within the budget. Score represents the aggregated task score.}
\label{tab: shopping}
\begin{tabular}{lccccc}
\toprule
Model & Score & PID & Qty\_Size & Voucher & Balance \\
\midrule
Gemini-3-Flash & 0.590 & 0.850 & 0.417 & 0.567 & 0.417 \\
Gemini-3-Pro   & \textbf{0.721} & 0.901 & \textbf{0.600} & 0.567 & \textbf{0.583} \\
Gemini-2.5-Pro         & 0.377 & 0.700 & 0.192 & 0.351 & 0.192 \\
GPT-5                  & 0.691 & \textbf{0.901} & 0.582 & \textbf{0.620} & 0.565 \\
GPT-5-Mini             & 0.473 & 0.750 & 0.267 & 0.517 & 0.233 \\
GPT-4.1                & 0.197 & 0.350 & 0.050 & 0.250 & 0.050 \\
GPT-4.1-Mini           & 0.183 & 0.383 & 0.050 & 0.167 & 0.033 \\
DeepSeek-V3.2-Thinking & 0.550 & 0.808 & 0.350 & 0.550 & 0.350 \\
DeepSeek-V3.2          & 0.417 & 0.792 & 0.150 & 0.417 & 0.150 \\
\bottomrule
\end{tabular}
\end{table}

\subsection{Cross-Metric Consistency Analysis}



\begin{figure}[t]
  \centering
   \includegraphics[width=\linewidth]{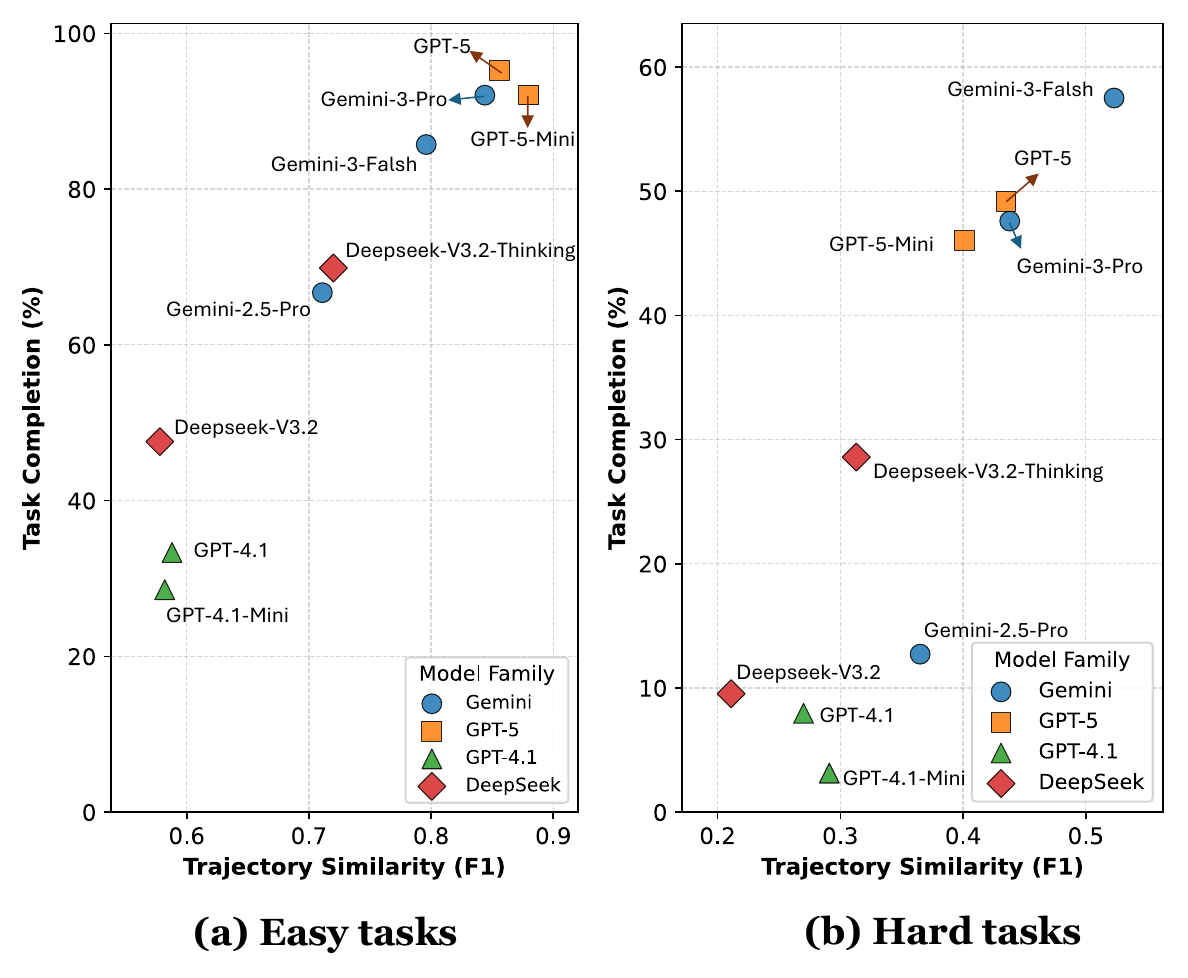}
  \caption{Alignment between task completion scores and trajectory similarity (F1), showing a strong positive correlation across evaluated models.}
  \label{fig: TC_F1}
  \Description{Side-by-side scatter plots comparing task completion and F1 on easy and hard tasks.}
\end{figure}

PAUSE adopts a multi-regime evaluation framework that combines target-grounded LLM judgment with trajectory-level, rule-based overlap metrics. The overlap metrics compute precision and recall over matched tool invocations, reflecting tool-use accuracy and coverage, with F1 summarizing overall trajectory similarity.
The two criteria capture complementary aspects of agent behavior and jointly improve evaluation reliability.
Figure~\ref{fig: TC_F1} plots the average task completion score from target-based LLM evaluation against the corresponding trajectory overlap F1 score for each model.
Across both easy and hard data \& log tracking tasks, we observe a strong positive correlation between task completion and trajectory overlap.
Models that achieve higher LLM-judged completion scores also exhibit greater overlap with reference trajectories, while weaker models underperform consistently under both measures.
This consistency indicates that LLM-based evaluation is behaviorally grounded, with the overlap metric providing an independent signal that corroborates its reliability.

\section{Error Analysis}
\subsection{Data \& Log Tracking}

\begin{figure}[t]
  \centering
  \includegraphics[width=0.9\linewidth]{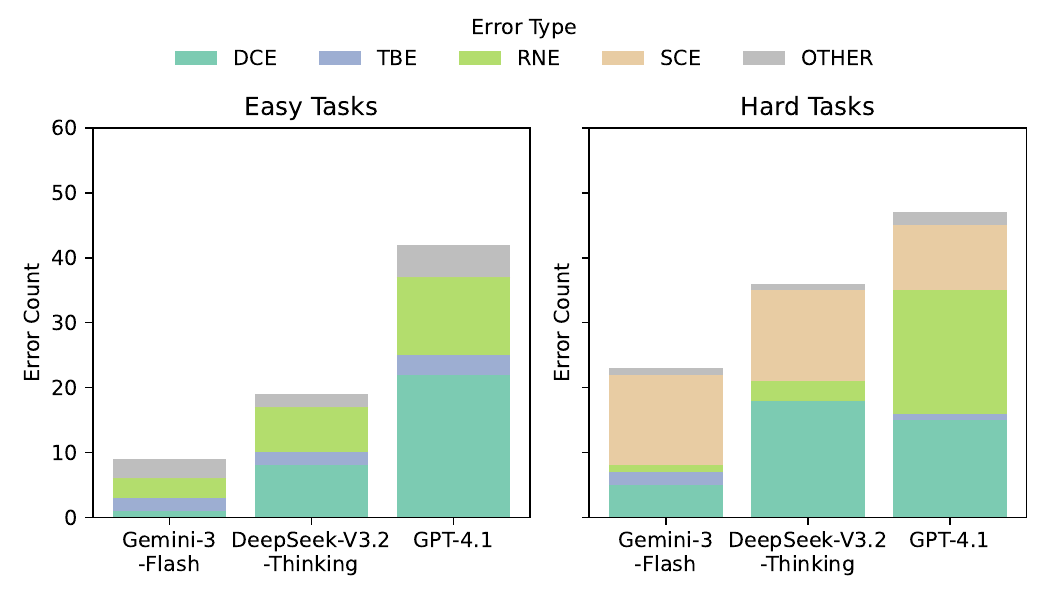}
  \caption{Error type distribution across three representative models on easy and hard tasks. Errors are grouped into DCE (data and computation error), TBE (tool bypassing error), RNE (resource navigation error), SCE (system configuration error), and OTHER.}
  \label{fig:error_distribution}
  \Description{None}
\end{figure}

To further investigate the challenges posed by system configuration reasoning, we leverage an LLM classifier to conduct a detailed error pattern analysis on three representative models: a strong proprietary model (Gemini-3-Flash), an open-source model (DeepSeek-V3.2-Thinking), and a weaker model (GPT-4.1). Errors are categorized into five types, covering data and computation issues, tool misuse or bypassing, resource navigation failures, system configuration reasoning errors, and others.

As shown in Figure~\ref{fig:error_distribution}, Gemini-3-Flash makes few mistakes on easy tasks. On hard tasks, its failures are concentrated primarily in system configuration reasoning, with only sporadic errors of other types. This implies that basic capabilities are no longer the bottleneck; instead, performance is constrained by limitations in system configuration and environment state reasoning.
In contrast, the two weaker models struggle not only with system configuration reasoning but also with more basic error patterns. DeepSeek-V3.2-Thinking exhibits a substantially higher number of errors across both easy and hard tasks, with a large portion of failures arising from data and computation issues. Closer inspection reveals that these errors are largely attributable to degraded temporal reasoning in long-context interactions: time-related information is frequently lost, resulting in incorrect tool inputs and subsequent task failures. GPT-4.1 performs worst overall, with frequent errors in both data/computation and resource navigation, suggesting that it lacks the fundamental capability to function effectively in a holistic, unified service environment.

We provide 2 representative error analyses to illustrate common failure modes observed across models.
Figure~\ref{fig:source_missing} shows a case where Gemini-3-Flash fails to recognize that partial data sources are not connected, leading to incomplete platform data and unsuccessful data extraction.
Figure~\ref{fig:resource_navigation} presents an example in which Gemini-2.5-Pro makes an incorrect resource navigation decision, resulting in failure to retrieve the intended data.

\begin{figure}[t]
\centering
\setlength{\fboxsep}{4pt}
\setlength{\fboxrule}{0.4pt}
\footnotesize
\fbox{%
\begin{minipage}{0.96\columnwidth}
\ttfamily
\raggedright

\textbf{Assistant model:} Gemini-3-Flash

\vspace{3pt}
\textbf{Task goal.}
The user asks the assistant to compare morning and evening runs over the past two weeks using sport records, summarize typical AZM and calories, check the latest care plan, reschedule an existing healthcare appointment with the same provider, and save a note explaining the comparison and appointment update.

\vspace{3pt}
\textbf{Expected behavior.}
After sport records are missing, the assistant should inspect source availability and identify that Xiaomi Mi Fitness is disconnected, which is the likely source of the missing recent sport records. It should ask the user to connect Xiaomi before performing the requested comparison.

\vspace{3pt}
\textbf{Key trajectory.}
\begin{itemize}
    \item get\_sport\_records(2024-03-21, 2024-04-04) $\rightarrow$ []
    \item get\_system\_settings() $\rightarrow$ fitbit: connected; xiaomi\_mi\_fitness: disconnected
    \item assistant asks the user to update Fitbit permissions/source state, but does not ask to connect Xiaomi
    \item get\_sport\_records(2024-03-21, 2024-04-04) $\rightarrow$ []
    \item assistant later uses outdated Fitbit records outside the requested two-week window
\end{itemize}

\vspace{3pt}
\textbf{Failure.}
The assistant fails to classify morning/evening runs and compare AZM/calories for the requested past-two-week period. The core error is incomplete multi-source awareness: it observes that Xiaomi is disconnected but does not use this information to recover missing sport records.

\vspace{3pt}
\textbf{Partial success.}
The assistant retrieves the care plan, cancels the existing appointment, creates a new appointment with the same provider, and saves a summary note.

\end{minipage}
}
\caption{Case summary of incomplete multi-source awareness, observed in Gemini-3-Flash rollouts.}
\Description{Case summary of incomplete multi-source awareness.}
\label{fig:source_missing}
\end{figure}

\begin{figure}[t]
\centering
\setlength{\fboxsep}{4pt}
\setlength{\fboxrule}{0.4pt}
\footnotesize
\fbox{%
\begin{minipage}{0.96\columnwidth}
\ttfamily
\raggedright

\textbf{Assistant model:} Gemini-2.5-Pro

\vspace{3pt}
\textbf{Task goal.}
The user asks the assistant to review 7 days of meal records, flag violations of vegan/lactose-free/gluten-free restrictions, correct three lunch entries, and compare daily calorie intake against a 2,300 kcal goal.

\vspace{3pt}
\textbf{Expected behavior.}
After updating the meal records, the assistant should retrieve or recompute meal-based calorie intake from the corrected records, then compare corrected days with other days in the same 7-day window.

\vspace{3pt}
\textbf{Key trajectory.}
\begin{itemize}
    \item get\_meal\_records(2024-02-08, 2024-02-14) $\rightarrow$ meal logs
    \item delete\_record(...) $\rightarrow$ removes three incorrect lunches
    \item create\_meal\_record(...) $\rightarrow$ creates corrected lunches
    \item get\_daily\_summary(2024-02-12/13/14) $\rightarrow$ calories.total
    \item assistant treats calories.total as calorie intake
\end{itemize}

\vspace{3pt}
\textbf{Failure.}
The assistant confuses calorie intake with calorie expenditure. It uses daily summary calories.total, which reflects calories burned, rather than summing calories from meal records. It also fails to retrieve the updated meal records after correction, so the reported intake values, goal status, and completion rates are incorrect.

\vspace{3pt}
\textbf{Partial success.}
The assistant identifies dietary conflicts, deletes the three incorrect meal records, and recreates corrected meals at similar timestamps.

\end{minipage}
}
\caption{Case summary of calorie burn misinterpreted as calorie intake, observed in Gemini-2.5-Pro rollouts.}
\Description{Case summary of calorie burn being incorrectly treated as calorie intake due to tool misuse.}
\label{fig:resource_navigation}
\end{figure}

Overall, the error analysis reveals a clear stratification in how models fail under our holistic, unified service environment. Strong models primarily break down at the level of system configuration reasoning when implicit constraints must be inferred, while weaker models fail more fundamentally, struggling to maintain basic data consistency and resource navigation.

\subsection{Ablation Study}

To further examine the impact of system configuration awareness on agent reasoning, we conduct an ablation study by augmenting the evaluation rollout with the policy guidance prompt $P_k^{\text{rollout}}$ used during reference trajectory generation.
This prompt explicitly specifies system configuration semantics and provides guidance on how agents should respond to missing permissions or authorization requirements during interaction.

\begin{figure}[t]
  \centering
  \includegraphics[width=0.9\linewidth]{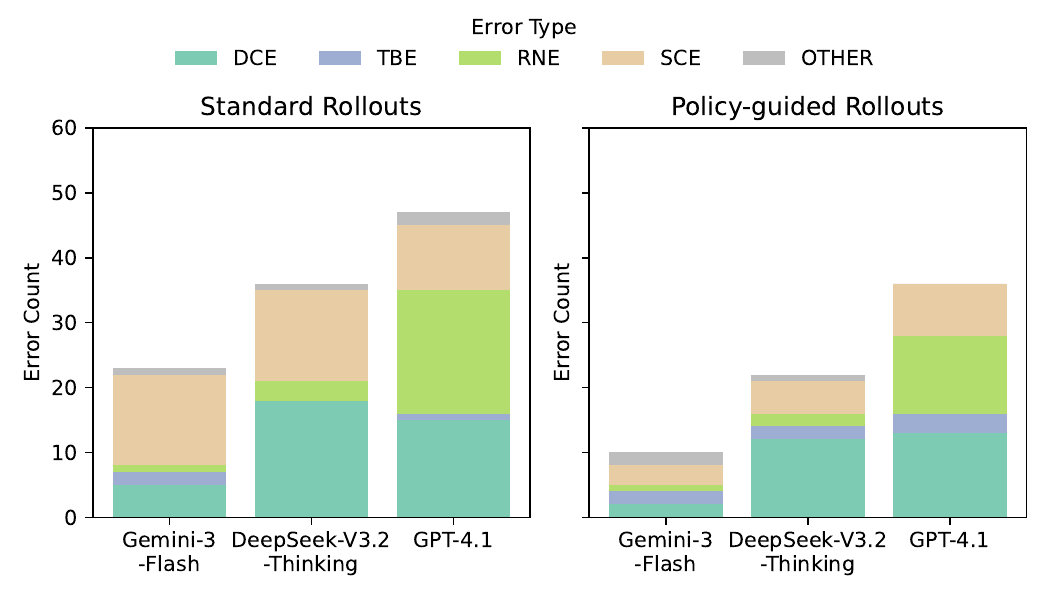}
  \caption{Error type distribution across three representative models on data \& long tracking (hard) tasks, with and without policy guidance $P_k^{\text{rollout}}$.
  Incorporating policy guidance consistently reduces system-related errors, with the most pronounced improvement observed for Gemini-3-Flash.}
  \Description{A comparison of error type distributions for three representative models on data and log tracking hard tasks, evaluated under two settings: with and without rollout policy guidance. The figure groups errors into several categories and shows that enabling policy guidance reduces system-related errors across models, with the largest reduction observed for Gemini-3-Flash.}

  \label{ablation}
\end{figure}

As shown in Figure~\ref{ablation}, incorporating policy guidance in data \& log tracking (hard) tasks leads to consistent performance improvements across all evaluated models.
In particular, Gemini-3-Flash benefits substantially from this ablation, exhibiting a remarkable reduction in system-related errors and a noticeable overall performance gain.
This suggests that stronger frontier models are able to effectively leverage explicit configuration policies to better align their reasoning with the underlying system constraints. DeepSeek-V3.2-Thinking continues to exhibit a moderate number of data computation and resource navigation errors but with reduced system configuration errors. GPT-4.1 shows only marginal improvement overall, with DCE and RNE errors remaining significant, and only a limited reduction observed in system configuration errors.

Overall, these findings underscore the importance of system configuration understanding for task execution in stateful service environments.
While explicit policy guidance improves robustness and reduces system-level failures, it does not fully close the performance gap across models, suggesting that deeper reasoning and grounding limitations cannot be addressed solely through prompt-level interventions.

\subsection{Shopping}
As reflected in Table~\ref{tab: shopping}, state-of-the-art proprietary models achieve strong overall performance on shopping tasks with high PID scores, indicating that they are generally capable of solving the underlying multi-constraint optimization problem in shopping tasks. However, their performance on Qty\_Size, Voucher, and Balance remains noticeably lower, suggesting that even strong models struggle to reliably handle stateful transitions such as quantity adjustment, voucher selection, and budget management across multi-step purchase workflows.

Notably, DeepSeek-V3.2-Thinking achieves competitive performance on shopping tasks, approaching proprietary models on several constraint-level metrics. This is likely because shopping tasks place limited demands on temporal reasoning, a known weakness observed in its data \& log tracking performance analysis.
In contrast, weaker models show consistently lower performance across most metrics, indicating limited effectiveness in handling complex constraints and maintaining consistent state transitions.

\subsection{LLM-Based and Human Evaluation Consistency}

To assess the reliability of LLM-as-judge evaluation, we compare the judgment results produced by different LLM evaluators against human annotations.
Specifically, we sample a subset of trajectories from both \emph{easy} and \emph{hard} tasks in data \& log tracking, and ask multiple LLMs to independently evaluate task completion.
We then compute the agreement rate between LLM judgments and human evaluations based on the achieved task targets.

\begin{table}[h]
    \centering
    \small
    \caption{Agreement rate between LLM-as-judge evaluation and human annotations across different evaluator models. Agreement is measured as the percentage of trajectories for which the LLM judgment matches the human evaluation.}
    \label{tab:llm_judge_consistency}
    \begin{tabular}{lcc}
        \toprule
        \textbf{Evaluator Model} & \textbf{Easy (\%)} & \textbf{Hard (\%)} \\
        \midrule
        GPT-5-mini        & 80.2 & 77.1 \\
        Gemini-3-Flash    & 86.1 & 84.7 \\
        Gemini-3-Pro      & 87.2 & 85.1 \\
        GPT-5             & 84.6 & 83.4 \\
        \bottomrule
    \end{tabular}
\end{table}

As shown in Table~\ref{tab:llm_judge_consistency}, LLM-as-judge evaluation exhibits a high level of consistency with human judgments across both easy and hard tasks.
Stronger models such as GPT-5 and Gemini-3-Pro achieve the highest agreement, indicating that LLM-based evaluation can serve as a reliable proxy for human assessment in complex, open-ended agent tasks. Notably, GPT-5-mini shows a relatively lower agreement rate, particularly on hard tasks. Closer inspection reveals that GPT-5-mini tends to assess task completion primarily based on the evaluated trajectory itself, rather than explicitly grounding its judgment in the provided reference trajectory summary.
This gap suggests that evaluator capability plays a critical role in maintaining evaluation reliability under long-context, multi-step contexts.

Overall, these results suggest that LLM-as-judge evaluation can serve as a practical and reliable alternative to human evaluation when sufficiently capable evaluator models are adopted.

\section{Conclusion}
In this work, we introduce PAUSE, a benchmark that evaluates personal service agents in a user-centric, holistic service environment. Our results indicate that state-of-the-art proprietary models demonstrate strong tool usage capabilities and resource navigation capabilities within the proposed unified personal service environment. However, they still exhibit consistent failure modes when reasoning over implicit system configurations, gated resources, and multi-step state transitions, suggesting that effective tool calling alone is insufficient for robust performance in realistic service settings. In contrast, weaker models struggle more fundamentally, often failing to navigate the expanded task space, misidentifying relevant resources, or breaking down under long-horizon interactions, highlighting a clear capability gap in both planning and resource navigation. Our extensive experiments and analysis demonstrate the efficacy of the proposed benchmark environment and provide deeper insights for agent applications under user-centric, holistic service environments.

In addition, PAUSE introduces an agentic pipeline for reliably generating validated and annotated tasks. This design enables scalable benchmark construction and provides infrastructure for further data distillation and model training.

\section{Limitations}
While we collect and evaluate multiple trajectories for a subset of experiments and observe generally stable performance trends across models, our evaluation does not systematically report pass\textasciicircum k and pass@k adopted in prior work \cite{yao2024tau, barres2025tau}, primarily due to practical time constraints. We plan to incorporate these metrics in future experiments. For open-ended tasks without canonical solution trajectories, our current annotation protocol does not explicitly enforce state verification over all key variables, which could further strengthen evaluation rigor. Nevertheless, for human-verified trajectories, state verification is largely consistent with target-based LLM evaluation, suggesting that the current protocol already provides reliable signals. Finally, while our study covers a diverse set of proprietary models, extending evaluation to a broader range of open-source models remains an important direction for future work.


\newpage

\bibliographystyle{ACM-Reference-Format}

\bibliography{references}

\appendix
\balance
\section{System Specification}
\label{sec:system_specification}
\subsection{System Design and Access Control}
PAUSE simulates an integrated health platform that aggregates user health and lifestyle data from heterogeneous sources, inspired by multi-source health data systems (e.g., \cite{FitbitDeveloper}, \cite{SpikeAPIDocs}, \cite{ValidicDocs}). 
Such platforms typically operate as an aggregation layer over wearable devices, third-party applications, and external health services. 
A key design principle is conditional data availability: user data is accessible only when the corresponding source is connected, authorized, and valid for the requested time range. 
Thus, aggregated summaries and historical records are partial and configuration-dependent rather than globally complete. 
This setting requires assistants to reason not only about the requested information, but also about whether the underlying data source and access state can support the request.

PAUSE models this through a layered platform design. 
The platform layer exposes unified, permission-checked resources, including aggregated wearable summaries, time-series statistics, sport records, meals, sessions, notes, reminders, and user profiles. 
External or sensitive resources, such as raw wearable data and healthcare records, are instead controlled by gated services. 
Their availability is exposed through feature-probing tools (e.g., get\_source\_features, get\_med\_features), which allow assistants to inspect whether specific capabilities are available under the current environment state. 
When prerequisites such as source authorization, subscription status, or provider enrollment are unmet, these tools return restricted statuses rather than exposing the corresponding functionality.

By modeling source connectivity, permissions, subscriptions, and gated services as persistent environment state, PAUSE evaluates whether assistants can identify valid access pathways, and maintain consistent stateful transitions across multi-step interactions. 
An assistant may fail not only by calling an incorrect tool, but also by overlooking hidden configuration constraints or assuming that unavailable data is globally accessible. 
Table~\ref{tab:tool_read_write_stats} and Table~\ref{tab:tool_inventory} summarize the tool inventory and read/write tool statistics in PAUSE.

\section{Few-shot Examples}
\label{sec:prompts_examples}
Figure~\ref{fig:fewshot_tasks_targets} presents representative task instructions paired with their corresponding target specifications. The examples are used as few-shot demonstrations for the task composer LLM in our user-centric task generation pipeline, guiding the translation from high-level user intent into structured, verifiable targets. During trajectory rollouts, these instructions are provided to a user agent, which issues task commands accordingly.

\begin{table}[H]
\centering
\small
\caption{Statistics of read and write tools in PAUSE, grouped by functional category.}
\begin{tabular}{lccc}
\toprule
\textbf{Category} & \textbf{Read Tools} & \textbf{Write Tools} & \textbf{Total} \\
\midrule
Medical Tools        & 3  & 3  & 6  \\
Platform Tools       & 18 & 11 & 29 \\
Shopping Tools       & 5  & 4  & 9  \\
Source Tools         & 5  & 1  & 6  \\
User Tools           & 0  & 7  & 7  \\
\midrule
\textbf{Overall}     & \textbf{31} & \textbf{26} & \textbf{57} \\
\bottomrule
\end{tabular}
\label{tab:tool_read_write_stats}
\end{table}

\begin{figure}[H]
\centering
\setlength{\fboxsep}{4pt}
\setlength{\fboxrule}{0.4pt}
\footnotesize
\fbox{%
\begin{minipage}{0.96\columnwidth}
\raggedright

\textbf{Example 1: Appointment Management with Activity Context}

\textbf{Label:} \texttt{appointment\_management\_with\_activity\_context\_note\_and\_reminder}

\textbf{Task Instruction (User Intent):}  
You are Samantha. You want to schedule a healthcare appointment without disrupting your usual workout routine
over the past two weeks. You first ask the assistant to review your recent activity patterns and recommend a suitable time.
After selecting an appropriate provider and appointment slot, you want the assistant to save a note with the appointment details
and set reminders so you do not forget.

\textbf{Targets:}
\begin{itemize}
    \item Infer the typical workout time window from sport records over the past two weeks.
    \item Identify available healthcare providers and appointment slots.
    \item Create an appointment that avoids the inferred workout window.
    \item Add a note containing the provider, appointment time, and avoidance rationale.
    \item Create reminder(s) leading up to the appointment date.
\end{itemize}

\textbf{Example 2: Evening Meal and Workout Pattern Analysis}

\textbf{Label:} \texttt{evening\_meal\_workout\_pattern\_with\_visualization\_and\_appointment}

\textbf{Task Instruction (User Intent):}  
You are Alex. You want to understand whether late-evening eating has been clashing with your evening workouts
and contributing to fatigue. Starting from last Wednesday, you ask the assistant to analyze food intake
and workout records, visualize calorie intake and expenditure trends, update your profile with one practical
evening preference, and help schedule a short healthcare appointment to discuss recovery concerns.

\textbf{Targets:}
\begin{itemize}
    \item Identify days with notable evening food intake since last Wednesday.
    \item Determine whether those days include evening workouts.
    \item Compare calorie intake and calories burned on those days.
    \item Visualize evening-related intake and expenditure trends.
    \item Update profile preferences based on the observed pattern.
    \item Create a healthcare appointment related to evening fatigue or recovery.
\end{itemize}

\textbf{Example 3: Weekly AZM Goal Analysis with Note and Reminder}

\textbf{Label:} \texttt{weekly\_goal\_day\_minute\_azm\_with\_note\_and\_reminder}

\textbf{Task Instruction (User Intent):}  
You are Samantha. You want to understand what helped you meet your Active Zone Minutes (AZM) goal in the past week.
You ask the assistant to identify successful days, zoom into a representative workout session,
visualize minute-level AZM accumulation, summarize the pattern in a note,
and create reminders to encourage repeating a similar routine in the coming week.

\textbf{Targets:}
\begin{itemize}
    \item Identify days in the past week that met the AZM goal and select one successful day.
    \item Retrieve sport records and select a primary workout session for that day.
    \item Retrieve and visualize minute-level AZM during the session window.
    \item Add a note summarizing the observed effort pattern.
    \item Recommend a suitable workout style consistent with the pattern.
    \item Create a daily reminder for the coming week.
\end{itemize}

\end{minipage}
}
\caption{Representative few-shot task instructions and target specifications used to condition the task composer LLM
in the user-centric task generation pipeline.}
\Description{Examples of few-shot task instructions paired with their corresponding target specifications.}
\label{fig:fewshot_tasks_targets}
\end{figure}

\begin{table*}[t]
\centering
\small
\caption{Tool inventory in PAUSE, grouped by functional category.}
\label{tab:tool_inventory}
\begin{tabular}{lll}
\toprule
\textbf{Category} & \textbf{Tool} & \textbf{Description and Access Constraints} \\
\midrule
\multirow{6}{*}{Medical Tools} 
& \texttt{med-get\_user\_profile} & Retrieve user medical profile \\
& \texttt{med-get\_provider\_list} & List available healthcare providers \\
& \texttt{med-get\_resources} & Query provider-specific medical resources \\
& \texttt{med-create\_appointment} & Create a healthcare appointment \\
& \texttt{med-cancel\_appointment} & Cancel an existing appointment \\
& \texttt{med-update\_user\_profile} & Update medical profile information \\
\midrule
\multirow{12}{*}{Platform Tools} 
& \texttt{get\_daily\_summary} & Daily aggregated wearable summary\\
& \texttt{get\_range\_summary} & Aggregated statistics over a date range \\
& \texttt{get\_hourly\_steps / mets / calories} & Hourly-level aggregated activity metrics \\
& \texttt{get\_sport\_records} & Retrieve sport and workout records \\
& \texttt{get\_session\_records} & Retrieve recorded activity sessions \\
& \texttt{get\_meal\_records} & Retrieve logged meal records \\
& \texttt{analysis\_meal} & Nutritional analysis over meal records \\
& \texttt{create\_session\_record} & Create a new activity session record \\
& \texttt{create\_meal\_record} & Create a meal record \\
& \texttt{delete\_record} & Delete platform-native records \\
& \texttt{list\_daily\_reminders / create / delete} & Reminder management \\
& \texttt{get\_user\_profile / update\_profile} & User profile access and update \\
\midrule
\multirow{9}{*}{Shopping Tools}
& \texttt{browse\_items} & Browse product catalog \\
& \texttt{inspect\_item} & Inspect item details (price, nutrition, discount) \\
& \texttt{add\_to\_cart / remove\_from\_cart} & Modify shopping cart state \\
& \texttt{get\_cart} & Retrieve current cart contents \\
& \texttt{get\_wallet} & Retrieve wallet balance and membership status \\
& \texttt{prepare\_order} & Prepare checkout order \\
& \texttt{authorize\_checkout} & Authorize payment and checkout \\
& \texttt{upgrade\_membership\_request} & Upgrade subscription tier \\
& \texttt{get\_transactions} & Retrieve transaction history \\
\midrule
\multirow{6}{*}{Source Tools}
& \texttt{get\_intraday\_steps / mets / calories} & Raw intraday wearable data\\
& \texttt{get\_intraday\_intensities} & Fine-grained activity intensity data \\
& \texttt{create\_activity\_plan} & Create source-level activity plans \\
& \texttt{get\_activity\_plan} & Retrieve activity plans \\
\midrule
\multirow{6}{*}{User Tools}
& \texttt{update\_source} & Connect wearable data sources \\
& \texttt{set\_raw\_data\_permission} & Grant access to raw wearable data \\
& \texttt{set\_user\_notes\_permission} & Grant access to personal notes \\
& \texttt{set\_purchase\_permission} & Grant purchase authorization \\
& \texttt{set\_med\_assistant\_permission} & Grant medical assistant permission \\
& \texttt{top\_up\_wallet} & Add balance to user wallet \\
\bottomrule
\end{tabular}
\end{table*}
\end{document}